\documentclass[letterpaper, 10 pt, conference]{ieeeconf}  

\IEEEoverridecommandlockouts                              

\usepackage{graphicx}
\usepackage{graphics} 
\usepackage{epsfig} 
\usepackage{mathptmx} 
\usepackage{times} 
\usepackage{amsmath} 
\usepackage{amssymb}  
\usepackage{booktabs}
\usepackage{multirow}
\usepackage{array}
\usepackage{tikz}
\usetikzlibrary{positioning,arrows.meta,calc}
\usepackage{cite}
\usepackage[hidelinks]{hyperref}
\usepackage{makecell}

\usepackage{xspace}
\newcommand{\etal}{et al.\xspace}

\title{\LARGE \bf
Learning Tactile-Aware Hand Representations from Human Hand–Object Interaction Data
}

\author{Yeonguk Yu, Jaehong Kim, and Byungok Han$^{*}$%
\thanks{The authors are with the Electronics and Telecommunications
Research Institute (ETRI), Republic of Korea
(e-mail: \{yeon\_guk, jhkim504, byungok.han\}@etri.re.kr).}%
\thanks{Corresponding author: Byungok Han.}%
}

\begin{document}
\bstctlcite{IEEEexample:BSTcontrol}

\maketitle
\thispagestyle{empty}
\pagestyle{empty}


\begin{abstract}

Tactile representation learning for dexterous manipulation often requires substantial robot-side data collection, limiting its scalability across different hands and sensors. We explore whether large-scale human hand-object interaction (HOI) data can provide a useful prior for robotic tactile learning. To this end, we propose \textit{HuTact}, a framework that pretrains tactile-aware hand representations from human HOI data without requiring real tactile signals during pretraining. HuTact transfers the interaction structure learned from human hand motions to robotic tactile observations and can be further adapted with a small amount of unlabeled robot interaction data. We evaluate HuTact on tactile perception and manipulation tasks, including grasp prediction, slip detection, object classification, and tactile-conditioned policy learning. Across these tasks, human HOI pretraining improves data efficiency and provides a strong initialization for downstream learning. Notably, with HOI pretraining followed by midtraining on unlabeled robot play data, HuTact achieves 88.26\% Macro-F1 on grasp prediction using only 25\% of the labeled downstream data, comparable to training from scratch on the full labeled dataset (88.23\%). These results suggest that human interaction data can serve as a scalable complementary source for robotic tactile representation learning.

\end{abstract}
\section{Introduction}
Tactile sensing enables humans to perceive and interact with the physical world, providing direct information about contact, force, texture, and motion that is difficult to obtain through vision alone~\cite{dahiya2009tactile}. 
Inspired by this remarkable human capability, tactile sensing has emerged as a critical modality for robotic manipulation, enabling robots to perceive subtle contact interactions, infer local physical properties, and regulate contact forces during complex manipulation tasks that demand precise and adaptive physical interaction~\cite{yuan2017gelsight, taylor2022gelslim, bhirangi2025anyskin}.

Accordingly, a wide range of tactile sensors have been developed to capture rich information arising from physical contact, including vision-based sensors (e.g., GelSight~\cite{yuan2017gelsight}, DIGIT~\cite{lambeta2020digit}) and magnetic-based sensors (e.g., AnySkin~\cite{bhirangi2025anyskin}, ReSkin~\cite{pmlr-v164-bhirangi22a}). 
Beyond advances in tactile sensing hardware, the growing availability of large-scale tactile interaction data has further motivated learning-based approaches to extract generalizable representations from raw tactile observations~\cite{yang2022touch}. In particular, recent works have explored self-supervised pretraining using large-scale tactile interaction data collected with respective tactile sensors, learning representations that can be transferred to downstream manipulation tasks~\cite{sharma2025selfsupervised, guzey2023dexterity, liu2024masked, fenganytouch2}.

Learning generalizable tactile representations can reduce the amount of task-specific data required for downstream tasks. For vision-based tactile sensing, Feng~\etal propose AnyTouch~\cite{fenganytouch}, which learns transferable representations from a sensor-aligned dataset collected across multiple visuo-tactile sensors. For non-vision-based tactile sensing, Sharma~\etal~\cite{sharma2025selfsupervised} apply DINO-based self-distillation to masked tactile sequences collected from a dexterous hand, while Guzey~\etal~\cite{guzey2023dexterity} pretrain tactile representations using task-agnostic robotic play data. However, existing approaches for non-vision-based tactile sensors still largely require real tactile interaction data collected directly with the target robotic hand and sensing hardware. As a result, adapting representation learning to a new robotic platform or tactile sensor requires additional data collection, introducing substantial time and cost for pretraining.

As an alternative, human hand-object interaction (HOI) data provide a promising source of pretraining supervision, where they contain large-scale and diverse interactions across objects, hand configurations, and manipulation behaviors~\cite{Chao_2021_CVPR,Yang_2022_CVPR,Hampali_2020_CVPR,zhan2024oakink2}. These datasets are particularly attractive for learning transferable hand representations because they provide (1) interaction patterns that can be shared across different hand embodiments, (2) large-scale interaction data that would be costly to collect on robotic platforms, and (3) an opportunity to reduce embodiment-specific data collection when adapting to a new robotic hand. Many of these datasets also provide temporally aligned 3D reconstructions of human hands and manipulated objects, allowing fingertip configurations and their evolution during interaction to be recovered without collecting data on a robotic platform. Although HOI datasets do not contain real tactile signals, their hand--object trajectories encode spatial and temporal structure associated with physical interaction. We hypothesize that this structure can serve as a transferable prior for robotic tactile representation learning and reduce the amount of robot-side tactile data required for adaptation.

In this work, our research question is: \textit{Can human hand-object interaction (HOI) data provide a transferable prior for robotic tactile representation learning?} To answer this question, we propose \textbf{HuTact}, a framework for learning tactile-aware robot hand representations from human HOI data. 
Rather than attempting to reconstruct real tactile signals from human data, HuTact learns the spatial and temporal structure of hand interactions from wrist-local fingertip trajectories.
To bridge human HOI pretraining and robotic tactile sensing, we separately tokenize fingertip coordinates and tactile observations, allowing the coordinate representation and shared Transformer backbone learned from HOI data to be retained while adapting the sensor-specific tactile input embedding. We further incorporate geometry-derived pseudo-contact signals as an auxiliary cue, encouraging the model to capture the relationship between fingertip coordinates and their corresponding tactile-like signals. This framework enables large-scale human interaction data to provide a transferable initialization for tactile-aware robot learning while reducing the amount of robot-side tactile data required for adaptation.

To validate the effectiveness and transferability of HuTact, we evaluate it on tactile-sensitive downstream tasks, including grasp prediction, slip detection, object classification, and policy learning on tactile-sensitive tasks, and analyze the factors that enable transfer from human HOI data to robotic tactile sensing. In summary, our contributions are as follows:
\begin{itemize}
\item We propose HuTact, a framework that uses large-scale human hand-object interaction data to pretrain tactile-aware robot hand representations without requiring real tactile signals during the initial pretraining stage.

\item We show that representations learned from HOI data transfer to robotic tactile perception and manipulation, improving grasp prediction, slip detection, object classification, and tactile-conditioned policy learning.

\item We provide systematic analyses of the transfer mechanism through architectural ablations and data-efficiency experiments, studying the roles of temporal modeling, token separation, pseudo-contact cues, and robot-side midtraining. Our results show that HOI pretraining can substantially reduce the amount of task-specific robotic tactile data required for downstream adaptation.
\end{itemize}
\section{Related Work}
\textbf{Tactile Sensing: }
Tactile sensors for robotic manipulation broadly include vision-based and non-optical sensing modalities. Vision-based sensors~\cite{yuan2017gelsight, lambeta2020digit} capture fine-grained contact geometry and have been applied to object identification~\cite{patel2020digger}, geometry sensing~\cite{dong2017improved,taylor2022gelslim}, pose estimation~\cite{villalonga2021tactile,bauza2023tac2pose}, and policy learning~\cite{xue2025reactive,chen2026implicitrdp}. However, their optical components can result in a relatively large form factor, making integration across multiple fingers challenging. Non-optical sensors instead use compact magnetic~\cite{bhirangi2025anyskin,uskin} or capacitive sensing elements~\cite{song2026fingertip} to enable distributed tactile sensing. In this work, we use the BrainCo Revo 2 hand~\cite{brainco_revo2}, which is equipped with fingertip capacitive sensors that measure normal force, tangential force, tangential-force direction, and proximity.

\textbf{Tactile Representation Learning: }
Recent work has explored transferable tactile representations through large-scale self-supervised pretraining. For vision-based tactile sensors, objectives developed for visual representation learning~\cite{he2022masked,oquab2023dinov2,radford2021learning} have been adapted to tactile observations~\cite{yang2024binding,liu2024masked,fenganytouch,fenganytouch2}. For skin-like tactile sensors, Sharma~\etal~\cite{sharma2025selfsupervised} apply self-distillation to masked tactile sequences from a dexterous hand, while Guzey~\etal~\cite{guzey2023dexterity} pretrain tactile representations using task-agnostic robotic play data. While these approaches demonstrate the benefits of tactile pretraining, they generally require interaction data collected with the target robot and tactile sensor. In contrast, HuTact initializes tactile-aware representations from human hand--object interaction data without real tactile signals. We pretrain on the spatial and temporal structure of wrist-local fingertip interactions, augmented with geometry-derived pseudo-contact cues, and subsequently adapt the representation to the target robotic tactile sensor.

\textbf{Robot Learning from Human Data: }
Human interaction data have been increasingly used to reduce robot-specific supervision for manipulation learning~\cite{hoque2026egodex,punamiya2026egoverse,beingbeyond2025beingh0,bi2025hrdt,wi2026tactalign}, by extracting hand poses, object motions, and interaction trajectories from large-scale demonstrations. HOI data have also been applied directly to dexterous manipulation: Zhao~\etal~\cite{zhao2025dexh2r} retarget human hand-object interactions to dexterous robot hands and learn a residual policy for embodiment mismatch, while Zeng~\etal~\cite{zeng2026flowhoi} use HOI data for vision-language-action learning. Large-scale datasets such as OakInk2~\cite{zhan2024oakink2}, TACO~\cite{liu2024taco}, ARCTIC~\cite{fan2023arctic}, and HOT3D~\cite{banerjee2025hot3d} provide scalable supervision for learning hand--object interaction structure. HuTact differs in objective: rather than transferring human motion directly to a robot policy, we use HOI data to pretrain a tactile-aware hand representation that is subsequently adapted to real robotic tactile sensing. To our knowledge, prior work has not systematically studied large-scale HOI data as a pretraining source for robotic tactile representation learning.

\section{Methodology}
HuTact is designed to transfer interaction structure learned from human HOI data to robotic tactile sensing. The key challenge is that the source HOI datasets provide hand and object geometry but no real tactile signals, whereas the target robot provides sensor-specific tactile observations. We address this mismatch by factorizing the input representation into two streams: a shared wrist-local fingertip coordinate stream and a sensor-dependent contact stream. The shared Transformer backbone is pretrained on human interactions, while the sensor-specific tactile input embedding is adapted when transferring to real robotic tactile observations.

\begin{figure*}[t]
    \centering
    \includegraphics[width=.90\linewidth]{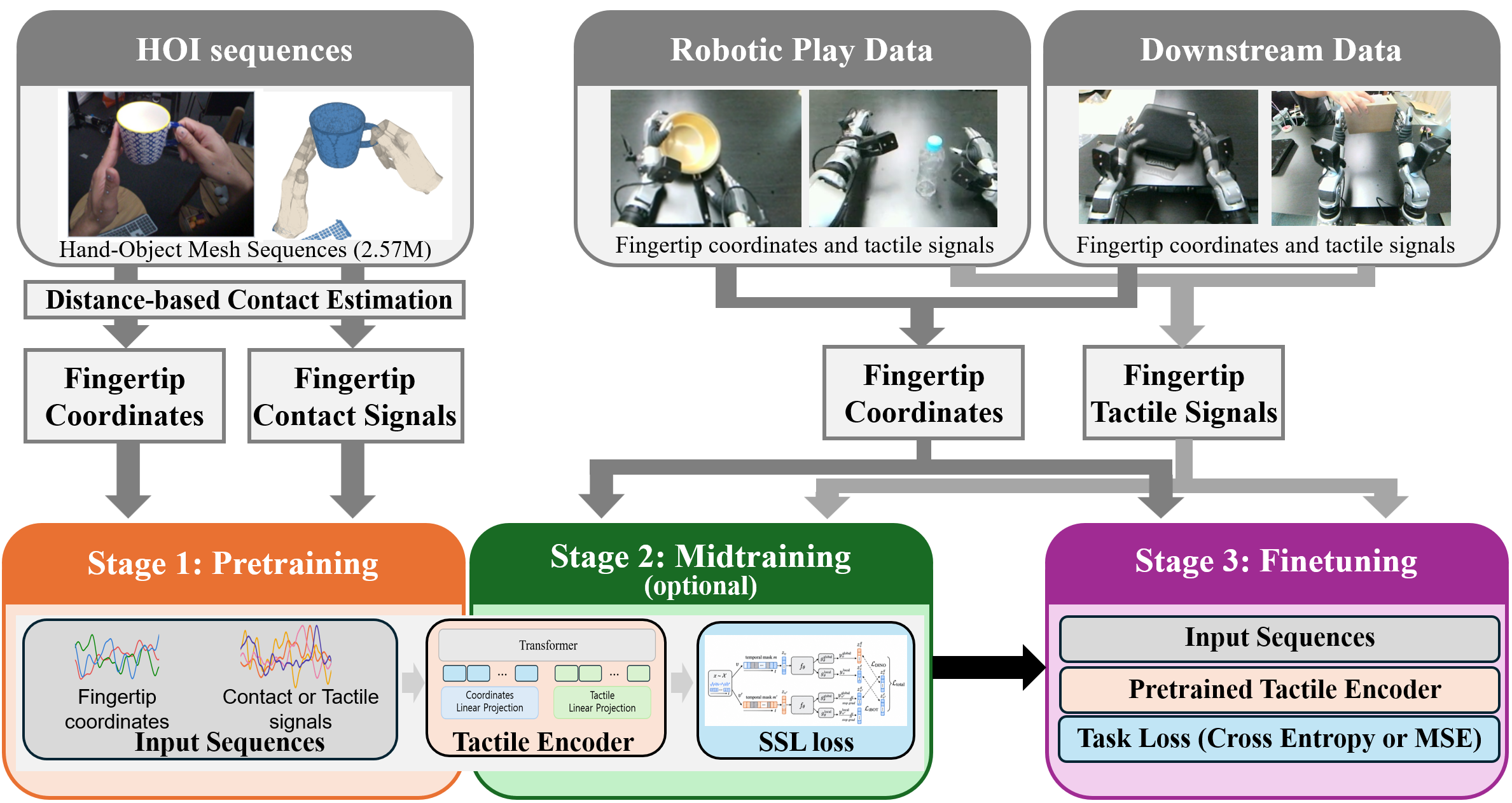}
    \caption{\textbf{Overview of HuTact.} Stage~1 pretrains the encoder on human HOI data using fingertip coordinates and geometry-derived pseudo-contact signals. Stage~2 optionally adapts the representation to real robotic tactile signals using unlabeled play data, and Stage~3 fine-tunes the encoder for downstream robotic tasks.}
    \label{fig:framework}
\end{figure*}

\subsection{Problem setup}

We consider two dexterous robot hands (BrainCo Revo2 Hand) with $N{=}10$ fingertip tactile sensors. At each timestep~$t$, sensor $i$ captures a signal $\mathbf{s}_i^t \in \mathbb{R}^{C}$ with $C{=}4$ channels---normal force, tangential force, tangential direction, and proximity---and forward kinematics provides the
corresponding fingertip coordinate $\mathbf{c}_i^t \in \mathbb{R}^{3}$ in the
wrist-local frame as in~\cite{bi2025hrdt}. The model operates on a temporal window
\begin{equation}
\mathbf{X} = \{(\mathbf{c}_i^t,\, \mathbf{s}_i^t)\}_{i=1,\dots,N}^{t=1,\dots,T},
\end{equation}
with $T{=}15$ frames sampled at $30$\,Hz, spanning $0.5$\,s of contact. Then, we pretrain a tactile encoder $f_\theta$ with human-object interaction (HOI) data and subsequently transfer it to robotic downstream tasks. To maintain a consistent input structure across the human and robot domains, human hand samples are represented using fingertip coordinates paired with a one-dimensional geometry-derived pseudo-contact signal (i.e., $C=1$), whereas robot observations use the corresponding fingertip coordinates paired with four-dimensional tactile signals. 

\subsection{HOI Data Processing}
The pretraining corpus is assembled from HOI dataset. We obtain its 3D coordinate from the MANO hand model~\cite{MANO:SIGGRAPHASIA:2017} and extract a one-dimensional pseudo-contact signal from the geometric relationship between the hand and object meshes. First, we convert the MANO hand representation~\cite{MANO:SIGGRAPHASIA:2017} into a common 21-joint skeleton by augmenting the 16 MANO kinematic joints with five fingertip vertices and reordering them into a unified joint convention. From this representation, we extract wrist-local fingertip coordinates and pseudo-contact signals for HOI pretraining.

\textbf{Fingertip coordinates.}
For each hand and frame, we construct a wrist-local coordinate frame from the hand geometry. The wrist defines the origin, while the palm-normal and finger directions are derived from the pinky--thumb and wrist--middle-finger axes and orthogonalized to form a right-handed basis. We transform all joints into this frame and retain the five fingertips per hand, yielding $N{=}10$ coordinates $\mathbf{c}_i^t\in\mathbb{R}^3$. During the experiment, robot fingertip coordinates from forward kinematics use the same local convention.

\textbf{Pseudo-contact signals.}
We define a surface patch $\mathcal{P}_i^t$ over each fingertip pad rather than estimating contact from a single fingertip point. MANO vertices are assigned to joints using dominant skinning weights, and each fingertip is augmented with the 20 nearest rest-pose vertices around its fingertip vertex. At each frame, we retain only vertices on the palmar side using the local pad direction, preventing dorsal-side vertices from producing spurious contact responses.

For each retained vertex $v \in \mathcal{P}_i^t$, we compute the shortest point-to-triangle distance $d_v^t$ to the object surface. Given a contact threshold of $d_{\max}=10$\,mm, we define
\begin{equation}
u_v^t =
\frac{\max(d_{\max}-d_v^t,\,0)}{d_{\max}},
\qquad
r_v^t = (u_v^t)^2(3-2u_v^t),
\end{equation}
where $r_v^t\in[0,1]$ provides a smooth proximity response.

We then aggregate the vertex responses using their associated surface areas:
\begin{equation}
s_i^t =
\frac{
\sum_{v\in\mathcal{P}_i^t} a_v^t r_v^t
}{
\sum_{v\in\mathcal{P}_i^t} a_v^t
},
\end{equation}
where $a_v^t$ is the representative surface area of vertex $v$, computed from its incident triangles. This area-weighted formulation reduces sensitivity to mesh tessellation and yields a bounded proximity-like signal $s_i^t\in[0,1]$, which increases as a larger portion of the fingertip pad approaches the object surface. When multiple objects are present, we select the object yielding the largest $s_i^t$ for each fingertip. For HOI pretraining, we use $s_i^t$ as a one-dimensional contact observation ($C=1$).

\subsection{Architecture}
\begin{figure}[t]
    \centering
    \includegraphics[width=1.0\linewidth]{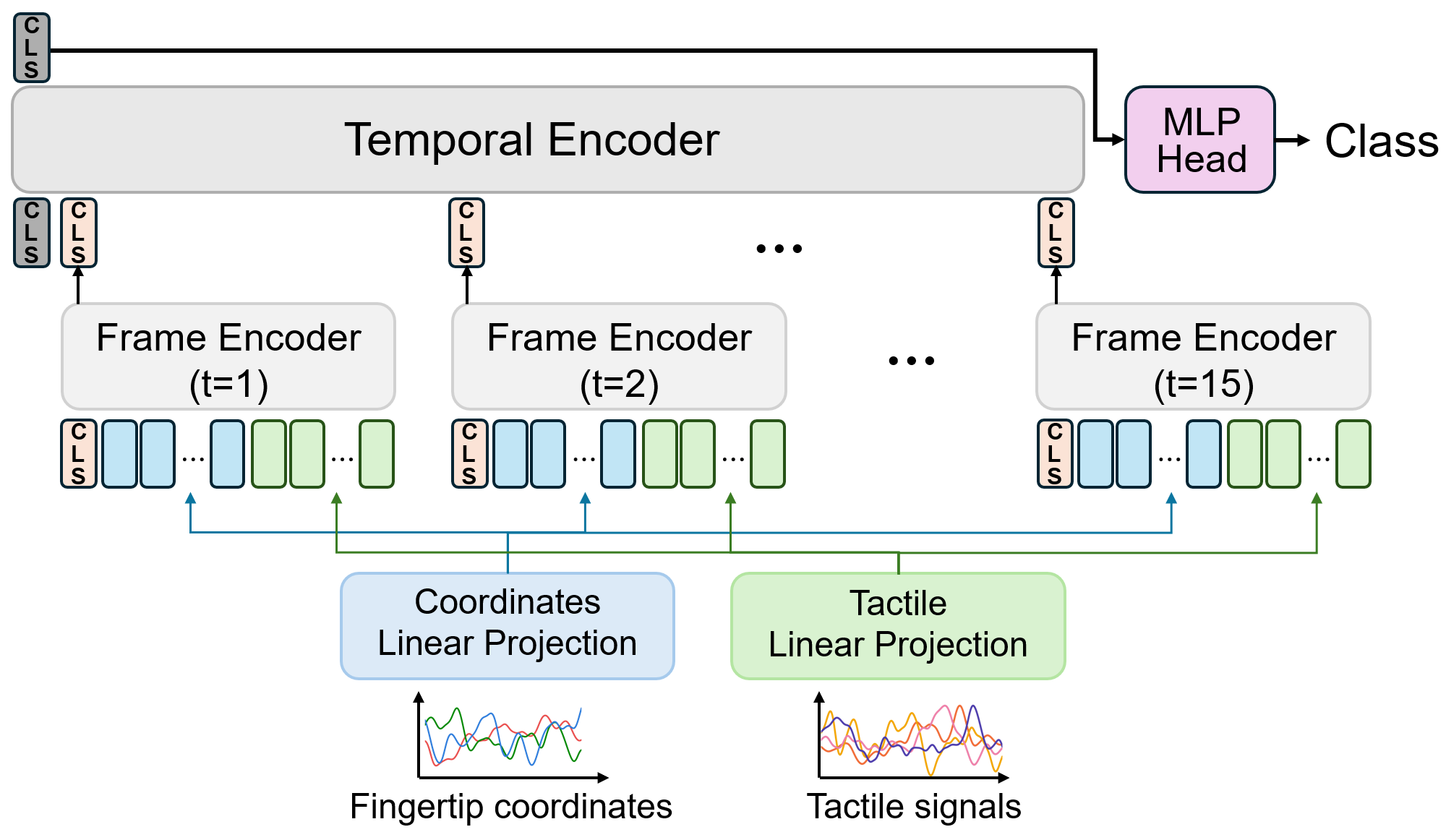}
    \caption{Illustration of backbone architecture. Each frame is encoded from separate tactile and wrist-local fingertip coordinate token streams. The resulting frame-level CLS tokens are then aggregated across 15 frames by a temporal Transformer to produce a window-level representation for downstream tasks.}
    \label{fig:arch}
\end{figure}

As shown in Fig.~\ref{fig:arch}, HuTact consists of separate coordinate and tactile tokenizers, a frame encoder, and a temporal encoder.

\textbf{Tokenizers. }Within a frame $t$, fingertip coordinates $c^t$ and tactile signals $s^t$ are tokenized as two separate streams: $\{\mathbf{c}_i^t\}$ and $\{\mathbf{s}_i^t\}$ are projected by distinct input embeddings into $N$ coordinate tokens and $N$ tactile tokens, each stream normalized with its own statistics. Specifically, the tokenizing process can be formalized as follows:
\begin{equation}
z_{c,i}^{t} = \mathrm{LN}_{c}\!\left(
\mathbf{W}_{c}\mathbf{c}^{t}_i+b_{c}
\right),
\qquad
z_{s,i}^{t} = \mathrm{LN}_{s}\!\left(
\mathbf{W}_{s}\mathbf{s}^{t}_i+b_{s}
\right), 
\label{eq:tokenizer}
\end{equation}
where the $z_{c,i}^t\in \mathbb{R}^D$ and $z_{s,i}^t\in \mathbb{R}^D$ denote the coordinate tokens and tactile tokens for the $i$-th finger, respectively. Also, $\mathbf{LN}$, $\mathbf{W}\in \mathbb{R}^{D\times m}$ ($m$ is the dimension of input), and $b\in \mathbb{R}^{D}$ refer to the Layer Norm~\cite{ba2016layer}, linear weight, and bias, respectively. Finally, we concatenate these tokens with \texttt{[CLS]} token as:
\begin{equation}
\mathbf{Z}^{t} = \left[
\mathbf{z}^t_{\mathrm{CLS}},
\{\mathbf{z}_{c,i}^{t}\}_{i=1}^{N},
\{\mathbf{z}_{s,i}^{t}\}_{i=1}^{N}
\right].
\label{eq:tokens}
\end{equation}
As shown in this Eq~\ref{eq:tokenizer} and Eq~\ref{eq:tokens}, we separate the token input for the tactile signals and fingertip coordinates unlike previous approaches~\cite{sharma2025selfsupervised}. This \textbf{token separation} is particularly important for transferring from HOI pretraining to real tactile sensing: while the contact signal used during pretraining and the real tactile signal have different dimensionalities (i.e., 1~dim vs. 4 dim), the coordinate tokenizer remains unchanged and can be directly reused after pretraining.

\textbf{Frame encoder.} 
Given the concatenated token sequence $\mathbf{Z}^t$ at each timestep $t \in \{1,2,\ldots,T\}$, we apply 2D multimodal RoPE~\cite{wang2024qwen2} using hand and finger identities as the two positional dimensions. We then process the tokens with a lightweight 4-layer Transformer encoder, following the remaining architectural hyperparameters of ViT-Tiny architecture~\cite{touvron2021deit}. Through self-attention, the encoder captures interactions between tactile observations and fingertip coordinates, producing a unified representation of the hand-object contact state at each timestep. The frame-level representations are then passed to the temporal encoder to model their relationship over time. Specifically, we take the \texttt{[CLS]} token from each frame encoder output and concatenate them along the token dimension to construct a sequence of frame-level representations. Given $T=15$ input frames, we obtain
\begin{equation}
\mathbf{H}
=
\left[
\mathbf{h}_{\mathrm{CLS}},
\hat{\mathbf{z}}^{1}_{\mathrm{CLS}},
\hat{\mathbf{z}}^{2}_{\mathrm{CLS}},
\ldots,
\hat{\mathbf{z}}^{15}_{\mathrm{CLS}}
\right],
\end{equation}
where $\mathbf{h}_{\mathrm{CLS}}$ and $\hat{\mathbf{z}}^{i}_{\mathrm{CLS}}$ are new CLS token for temporal encoder and output CLS token for $i$-th frame with frame encoder, respectively.

\textbf{Temporal encoder.}
Given the frame-level sequence $\mathbf{H}$, the temporal encoder applies RoPE based on the absolute frame index and models temporal dependencies across the $T=15$ frame-level \texttt{[CLS]} tokens. It uses the same Transformer architecture as the frame encoder with independent parameters. The output temporal \texttt{[CLS]} token $\hat{\mathbf{h}}_{\mathrm{CLS}}$ is used as the final window-level representation. This representation summarizes the spatiotemporal hand-object interaction over the entire $T=15$ frame window and is subsequently used by the downstream prediction head.


\subsection{Training objectives and stages}
\textbf{Objective.}
We pretrain the model using temporal self-distillation following the DINOv2~\cite{oquab2023dinov2} as~\cite{sharma2025selfsupervised} in the pretraining and midtraining stages. A student network is trained to match an exponential-moving-average (EMA) teacher by combining a distillation loss on the temporal \texttt{[CLS]} embedding with a masked-token prediction loss on the frame-level tokens.

\textbf{Stage 1: Pretraining on HOI data.}
In the first stage, the encoder is pretrained for 200 epochs using the objective described above on the HOI data. The data consists of contact signals and fingertip coordinates derived from public HOI data. Note that, no robot data are used at any point during this stage.

\begin{figure*}[!t]
    \centering
    \includegraphics[width=.98\linewidth]{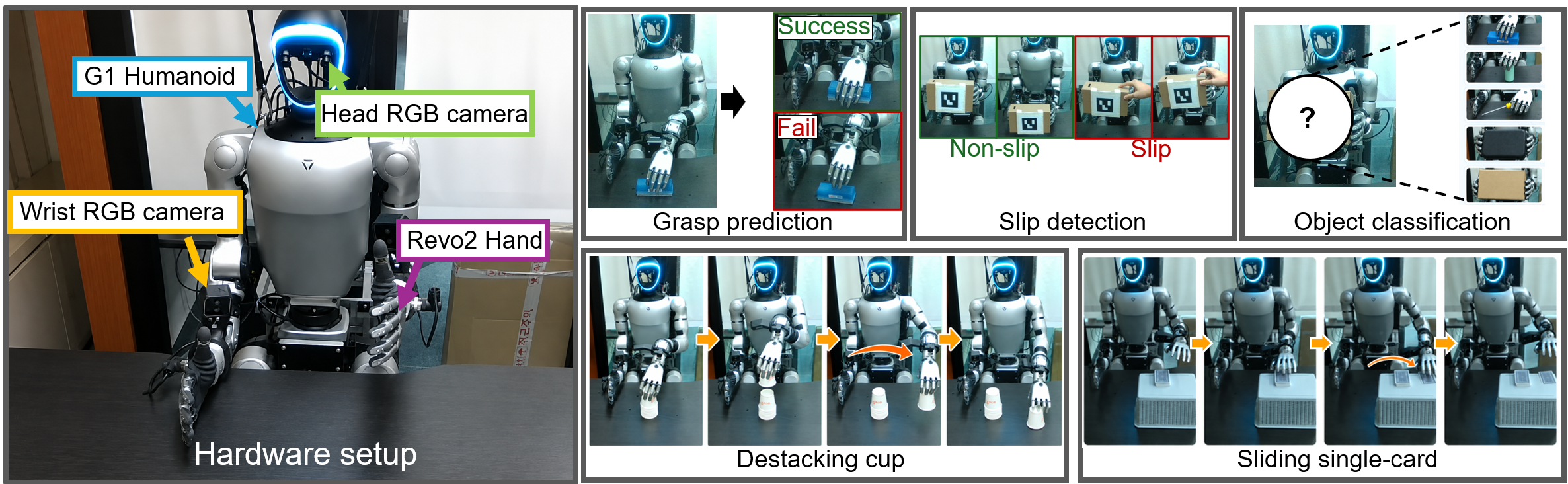}
    \caption{Hardware setup and downstream tasks. We use a Unitree G1 humanoid equipped with a BrainCo Revo2 dexterous hand. Downstream tasks includes grasp prediction, slip detection, object classification, cup destacking, and single-card sliding.}
    \label{fig:setup}
\end{figure*}

\textbf{Stage 2 (Optional): Midtraining on robotic play data.}
In the second stage, we optionally continue self-supervised training using the same objective and optimization recipe on a small unlabeled robotic play data. Because the real robot provides $C=4$ tactile sensor channels, whereas Stage~1 uses a different pseudo-tactile channel configuration, we re-initialize the tactile input tokenizer at the transition between the two stages while retaining all remaining pretrained parameters. Each epoch repeats the robot corpus 32 times, preserving approximately the same number of training samples per epoch as in Stage~1 ($\approx505$k samples per epoch). We train for 100 epochs using the same warmup-cosine learning-rate schedule. Hand poses are min-max normalized to the unit box during this stage.

\textbf{Stage 3: Downstream transfer.}
For downstream transfer, we fine-tune the full encoder jointly with an attentive-pooling task head. When Stage 2 midtraining is not conducted, we re-initialize the tactile input tokenizer before downstream fine-tuning; otherwise, the tokenizer adapted during Stage 2 is used. For the given downstream data, we use AdamW with a constant learning rate of $1\times10^{-4}$ for 100 epochs.

\textbf{Optimization.}
We use AdamW with a peak learning rate of $1\times10^{-4}$ and a batch size of 512. The learning rate is linearly warmed up for the first 10 epochs and subsequently decayed using a cosine schedule to $1\times10^{-6}$. The teacher temperature is increased from 0.04 to 0.07 during the first 2 epochs. The EMA momentum of the teacher is increased from 0.994 to 1.0 throughout training, while the weight decay follows a cosine schedule ending at 0.04.

\section{Experiments}
\subsection{Experimental setup}

We use a Unitree G1 humanoid equipped with a BrainCo Revo2 dexterous hand~\cite{brainco_revo2}, which has capacitive tactile sensors at each fingertip, as shown in Fig.~\ref{fig:setup}.

\textbf{Pretraining datasets.}
For HOI pretraining, we use HOT3D~\cite{banerjee2025hot3d}, OakInk-v2~\cite{zhan2024oakink2}, TACO~\cite{liu2024taco}, and ARCTIC~\cite{fan2023arctic}, extracting fingertip coordinates and pseudo-contact signals from 2.57M frames ($\sim$24 hours). We additionally collect 80K frames ($\sim$44 minutes) of Revo2 play data containing fingertip coordinates and tactile signals through Meta Quest~3 teleoperation~\cite{xr-teleoperate} with six objects (e.g., plastic bottle, bubble wrap, driver, pot, sponge, and towel).

\textbf{Downstream datasets.}
For \emph{grasp prediction}, we collect 590 teleoperated grasp episodes over five objects, totaling 18K frames ($\sim$10 minutes), with each episode labeled by grasp success. For \emph{slip detection}, we hold the robot hand fixed and manually move the grasped object while tracking an attached marker; marker motion is labeled as slip, yielding 18 episodes over six object conditions and 51K frames ($\sim$28 minutes). For \emph{object classification}, we reuse successful grasp episodes, extract the result-phase frames, and label the grasped object identity, yielding 252 episodes and 43K frames ($\sim$24 minutes).

For ID evaluation, we perform episode-level cross-validation before constructing temporal windows, ensuring that all windows from the same episode remain in the same split. We use stratified 4-fold cross-validation for grasp prediction and object classification, with 75\% of the episodes for training and 25\% for evaluation in each fold, and 3-fold cross-validation for slip detection. Reported ID performance is averaged across the corresponding folds.

For policy learning, we collect two datasets using a scripted controller. The cup-grasping and card-sliding datasets contain 40 and 55 episodes, respectively, each totaling approximately 30K frames ($\sim$17 minutes). For each policy configuration, success rates are computed over 40 evaluation trials per task.

\subsection{Performance on downstream tasks}
\begin{table*}[t]
\centering
\caption{Comparison of pretraining strategies on downstream perception tasks. Macro-F1 scores are mean $\pm$ standard deviation over 9 runs. ``Last'' denotes the final epoch and ``Avg'' the mean over evaluations every 10 epochs.}
\label{tab:downstream}
\resizebox{0.8\textwidth}{!}{
\begin{tabular}{llcccc}
\toprule
\textbf{Task} & \textbf{Method} 
& \textbf{ID Last} 
& \textbf{ID Avg} 
& \textbf{OOD Last} 
& \textbf{OOD Avg} \\
\midrule

\multirow{4}{*}{Grasp Prediction}
& From Scratch
& 88.23 $\pm$ 0.78
& 86.39 $\pm$ 0.38
& 74.92 $\pm$ 9.78
& 74.12 $\pm$ 9.78 \\

& Revo2 Pretraining
& 92.15 $\pm$ 0.40
& 92.09 $\pm$ 0.35
& 77.84 $\pm$ 9.31
& 77.51 $\pm$ 9.15 \\

& HOI Pretraining (Ours)
& \underline{92.61} $\pm$ 0.74
& \underline{92.18} $\pm$ 0.75
& \underline{79.21} $\pm$ 10.06
& \underline{78.46} $\pm$ 10.05 \\

& HOI + Revo2 Midtraining (Ours)
& \textbf{93.29 $\pm$ 0.53}
& \textbf{93.24 $\pm$ 0.54}
& \textbf{80.27 $\pm$ 9.52}
& \textbf{80.02 $\pm$ 9.49} \\
\midrule

\multirow{4}{*}{Slip Detection}
& From Scratch
& \underline{75.68} $\pm$ 1.73
& 73.14 $\pm$ 0.73
& 69.28 $\pm$ 4.55
& 67.39 $\pm$ 3.24 \\

& Revo2 Pretraining
& 72.37 $\pm$ 1.69
& 71.55 $\pm$ 1.96
& 65.75 $\pm$ 4.38
& 65.80 $\pm$ 3.87 \\

& HOI Pretraining (Ours)
& 75.28 $\pm$ 1.00
& \underline{73.78} $\pm$ 0.80
& \underline{69.56} $\pm$ 4.38
& \underline{68.86} $\pm$ 3.14 \\

& HOI + Revo2 Midtraining (Ours)
& \textbf{76.88 $\pm$ 1.03}
& \textbf{76.41 $\pm$ 0.85}
& \textbf{69.79 $\pm$ 4.24}
& \textbf{69.83 $\pm$ 3.75} \\
\midrule

\multirow{4}{*}{Object Classification}
& From Scratch
& 85.11 $\pm$ 1.35
& 82.96 $\pm$ 0.71
& N/A
& N/A \\

& Revo2 Pretraining
& \underline{89.07} $\pm$ 1.31
& \underline{88.85} $\pm$ 1.23
& N/A
& N/A \\

& HOI Pretraining (Ours)
& 87.64 $\pm$ 0.82
& 87.31 $\pm$ 1.08
& N/A
& N/A \\

& HOI + Revo2 Midtraining (Ours)
& \textbf{89.89 $\pm$ 1.52}
& \textbf{89.70 $\pm$ 1.48}
& N/A
& N/A \\
\bottomrule
\end{tabular}
}
\end{table*}

\begin{table*}[t]
\centering
\caption{Success rates (\%) on tactile-sensitive manipulation tasks.}
\label{tab:policy_results}
\resizebox{0.7\linewidth}{!}{
\begin{tabular}{llccc}
\toprule
\textbf{Model} & \textbf{Tactile Encoder}
& \textbf{Cup Destacking}
& \textbf{Single-Card Sliding}
& \textbf{Avg.} \\
\midrule

\multirow{3}{*}{ACT}
& None
& 30.0 & 37.5 & 33.8 \\

& From Scratch
& 5.0 & 60.0 & 32.5 \\

& HOI + Revo2 Midtraining (Ours)
& \textbf{37.5} & \textbf{62.5} & \textbf{50.0} \\
\midrule

\multirow{3}{*}{Diffusion Policy}
& None
& 27.5 & 17.5 & 22.5 \\

& From Scratch
& 37.5 & 40.0 & 38.8 \\

& HOI + Revo2 Midtraining (Ours)
& \textbf{47.5} & \textbf{52.5} & \textbf{50.0} \\
\bottomrule
\end{tabular}
}
\end{table*}

In Table~\ref{tab:downstream}, we evaluate HuTact on three downstream perception tasks: grasp prediction, slip detection, and object classification. Grasp prediction and slip detection are binary classification tasks that predict grasp success from observations collected 1 second after grasp execution and whether the current grasp is slipping or stable, respectively. Object classification is a five-way classification task that predicts the identity of the grasped object from observations collected during the post-grasp phase. Also, in the ID setting, all object categories are included in both the training and evaluation sets. In the OOD setting, we use a leave-one-object-out protocol, where one object is excluded from training and used only for evaluation, while the remaining objects are used exclusively for training (except object classification). We compare HuTact against two baselines: (i) \emph{From Scratch}, where the model is randomly initialized, and (ii) \emph{Revo2 Pretraining}, where the encoder is pretrained on Revo2 play data using the same self-supervised objective as Stage~1. 

HOI pretraining (Stage 1 only) consistently improves grasp prediction over both training from scratch and Revo2-only pretraining, with particularly clear gains in the OOD setting. For slip detection, HOI pretraining provides smaller but consistent gains in ID Avg and OOD performance, while achieving comparable ID Last performance to training from scratch. Revo2-only pretraining performs worse on slip detection, indicating that a small amount of robot play data alone is insufficient to learn transferable features for this task. For object classification, however, Revo2 pretraining outperforms HOI pretraining, suggesting that real tactile observations contain additional object-specific information. Across all three tasks, HOI pretraining followed by Revo2 midtraining achieves the best overall performance, demonstrating that large-scale HOI data and robot-side tactile data provide complementary benefits for downstream tactile representation learning.

\subsection{Performance with action policy}
In Table~\ref{tab:policy_results}, we evaluate the learned tactile representation on two contact-sensitive manipulation tasks using ACT~\cite{Zhao-RSS-23} and Diffusion Policy (DP)~\cite{chi2025diffusion}. \emph{Cup Destacking} requires the robot to separate a cup from a stack while maintaining a stable grasp, whereas \emph{Single-Card Sliding} requires the robot to slide a single card without displacing the remaining cards. We compare three tactile configurations: (i) no tactile input, (ii) a tactile encoder trained from scratch, and (iii) our HOI-pretrained encoder followed by Revo2 midtraining. We modify ResNet18-based ACT and DP with our tactile encoder. For ACT, the tactile representation is linearly projected to the Transformer embedding dimension and appended as a single token to the encoder input sequence. The tactile representation is used only as a conditioning input to the policy and is not provided to the VAE encoder. For Diffusion Policy, we concatenate the tactile representation directly with the visual observation features in the global conditioning vector.

Our tactile representation achieves the best overall performance for both policy architectures. With ACT, HOI + Revo2 midtraining improves the average success rate to 50.0\%, compared with 33.8\% without tactile input and 32.5\% with a tactile encoder trained from scratch. With Diffusion Policy, the improvement is more consistent across both tasks: our method achieves 47.5\% on Cup Destacking and 52.5\% on Single-Card Sliding, yielding an average success rate of 50.0\%, compared with 22.5\% without tactile input and 38.8\% from scratch. We observe that ACT with a tactile encoder trained from scratch often grasps multiple cups during Cup Destacking, which is counted as a failure and contributes to its low success rate of 5.0\%. These results indicate that the pretrained tactile representation provides useful contact-sensitive features for downstream policy learning beyond either omitting tactile input or learning the tactile encoder solely from task-specific demonstrations.

\subsection{Transfer performance comparison}
\begin{table}[t]
\centering
\caption{Cross-sensor transfer performance on downstream tasks.}
\label{tab:cross_embodiment}

\setlength{\tabcolsep}{2pt}
\renewcommand{\arraystretch}{1.0}

\begin{tabular}{@{}l *{6}{c} ccc@{}}
\toprule
&
\multicolumn{3}{c}{
  \makecell{\textbf{Encoder}\\\textbf{Architecture}}}
&
\multicolumn{3}{c}{
  \makecell{\textbf{Pretraining}\\\textbf{Source}}}
&
\makecell{\textbf{Grasp}\\\textbf{Pred.}}
&
\makecell{\textbf{Slip}\\\textbf{Det.}}
&
\makecell{\textbf{Object}\\\textbf{Cls.}}
\\
\midrule

& \multicolumn{3}{c}{Sparsh-Skin~\cite{sharma2025selfsupervised}}
& \multicolumn{3}{c}{Allegro + Xela uSkin}
& 87.8 & 70.6 & 81.0 \\

& \multicolumn{3}{c}{HuTact (Ours)}
& \multicolumn{3}{c}{HOI}
& \textbf{92.2} & \textbf{73.8} & \textbf{87.3} \\
\bottomrule
\end{tabular}
\end{table}

Table~\ref{tab:cross_embodiment} compares HuTact with Sparsh-Skin~\cite{sharma2025selfsupervised} to evaluate cross-sensor transfer across different pretraining sources and encoder architectures. We initialize Sparsh-Skin from the authors' publicly released checkpoint, pretrained on an Allegro hand equipped with Xela uSkin sensors, and fine-tune it on the Revo2 downstream tasks. Because the two platforms differ in the number of sensors and tactile signal dimensions, we reinitialize the patch and positional embeddings while retaining the remaining pretrained parameters. HuTact consistently outperforms Sparsh-Skin across all three downstream tasks. This performance gap is consistent with the benefits of token separation for transfer across sensing configurations and temporal modeling for capturing contact dynamics over time. These results indicate that HuTact provides an effective initialization for robotic tactile representation learning.

\subsection{Ablation Studies}
\begin{table}[t]
\centering
\caption{Ablation studies on in-distribution downstream tasks. Macro-F1 (\%) is averaged every 10 epochs from 10 to 100.}
\label{tab:ablation}

\setlength{\tabcolsep}{2pt}
\renewcommand{\arraystretch}{1.0}

\begin{tabular}{@{}l *{6}{c} ccc@{}}
\toprule
\multicolumn{10}{@{}l}{\textbf{(a) Architecture ablation}} \\
\midrule
&
\multicolumn{2}{c}{
  \makecell{\textbf{Temp.}\\\textbf{Encoder}}}
&
\multicolumn{2}{c}{
  \makecell{\textbf{Token}\\\textbf{Sep.}}}
&
\multicolumn{2}{c}{
  \makecell{\textbf{HOI}\\\textbf{Pretrain.}}}
&
\makecell{\textbf{Grasp}\\\textbf{Pred.}}
&
\makecell{\textbf{Slip}\\\textbf{Det.}}
&
\makecell{\textbf{Object}\\\textbf{Cls.}}
\\
\midrule

A
& \multicolumn{2}{c}{--}
& \multicolumn{2}{c}{--}
& \multicolumn{2}{c}{--}
& 84.8 & 60.8 & 81.8 \\

B
& \multicolumn{2}{c}{--}
& \multicolumn{2}{c}{--}
& \multicolumn{2}{c}{$\checkmark$}
& 87.7 & 60.0 & 83.7 \\

C
& \multicolumn{2}{c}{--}
& \multicolumn{2}{c}{$\checkmark$}
& \multicolumn{2}{c}{--}
& 84.3 & 61.8 & 83.7 \\

D
& \multicolumn{2}{c}{--}
& \multicolumn{2}{c}{$\checkmark$}
& \multicolumn{2}{c}{$\checkmark$}
& 91.0 & 60.1 & \textbf{87.9} \\

E
& \multicolumn{2}{c}{$\checkmark$}
& \multicolumn{2}{c}{$\checkmark$}
& \multicolumn{2}{c}{--}
& 86.4 & 73.1 & 83.0 \\

F
& \multicolumn{2}{c}{$\checkmark$}
& \multicolumn{2}{c}{$\checkmark$}
& \multicolumn{2}{c}{$\checkmark$}
& \textbf{92.2} & \textbf{73.8} & 87.3 \\

\midrule
\multicolumn{10}{@{}l}{\textbf{(b) Pseudo-contact ablation}} \\
\midrule
&
\multicolumn{3}{c}{
  \makecell{\textbf{Pseudo-Contact}\\\textbf{Signals}}}
&
\multicolumn{3}{c}{
  \makecell{\textbf{Revo2}\\\textbf{Midtrain.}}}
&
\makecell{\textbf{Grasp}\\\textbf{Pred.}}
&
\makecell{\textbf{Slip}\\\textbf{Det.}}
&
\makecell{\textbf{Object}\\\textbf{Cls.}}
\\
\midrule

F$^{\prime}$
& \multicolumn{3}{c}{--}
& \multicolumn{3}{c}{--}
& 92.0 & 72.6 & 87.3 \\

F
& \multicolumn{3}{c}{$\checkmark$}
& \multicolumn{3}{c}{--}
& 92.2 & 73.8 & 87.3 \\

G$^{\prime}$
& \multicolumn{3}{c}{--}
& \multicolumn{3}{c}{$\checkmark$}
& 92.9 & 73.0 & 89.5 \\

G
& \multicolumn{3}{c}{$\checkmark$}
& \multicolumn{3}{c}{$\checkmark$}
& \textbf{93.2} & \textbf{76.4} & \textbf{89.7} \\
\bottomrule
\end{tabular}
\end{table}

In Table~\ref{tab:ablation}-(a), we analyze the effects of the temporal encoder, token separation, and HOI pretraining on the three downstream tasks. Configurations A and B use a non-temporal, unified-token architecture similar to prior tactile representation learning approaches~\cite{sharma2025selfsupervised}, serving as a reference for our architectural modifications. \textbf{(i)} The temporal encoder provides the largest gain for slip detection: without HOI pretraining, adding temporal modeling improves Macro-F1 from 61.8\% to 73.1\% (C$\rightarrow$E), highlighting the importance of modeling contact dynamics over time. \textbf{(ii)} Token separation is particularly effective when combined with HOI pretraining, improving grasp prediction from 87.7\% to 91.0\% and object classification from 83.7\% to 87.9\% (B$\rightarrow$D). Separating fingertip coordinates from sensor-specific tactile inputs also allows the coordinate tokenizer learned from HOI data to be retained during transfer to the robot domain, and HOI pretraining yields larger gains when separate tokenizers are used. (\textbf{iii}) The complete configuration (F) achieves the best grasp prediction and slip detection performance among the architecture variants.

Table~\ref{tab:ablation}-(b) examines the contribution of pseudo-contact signals during Stage~1 pretraining. \textbf{(i)} Pseudo-contact pretraining provides its clearest benefit on slip detection, a task that relies strongly on tactile contact information, improving Macro-F1 from 72.6\% to 73.8\% without Revo2 midtraining (F$^{\prime}\rightarrow$F), while the gains in grasp prediction and object classification are negligible. \textbf{(ii)} This benefit becomes more evident after Revo2 midtraining, where pseudo-contact improves slip detection from 73.0\% to 76.4\% (G$^{\prime}\rightarrow$G). These results suggest that pseudo-contact pretraining helps the model capture the relationship between fingertip-coordinate and tactile tokens, providing a more transferable initialization when adapting to real robotic tactile signals.

\subsection{Downstream Data Efficiency}
\begin{figure}[t]
    \centering
    \includegraphics[width=1.0\linewidth]{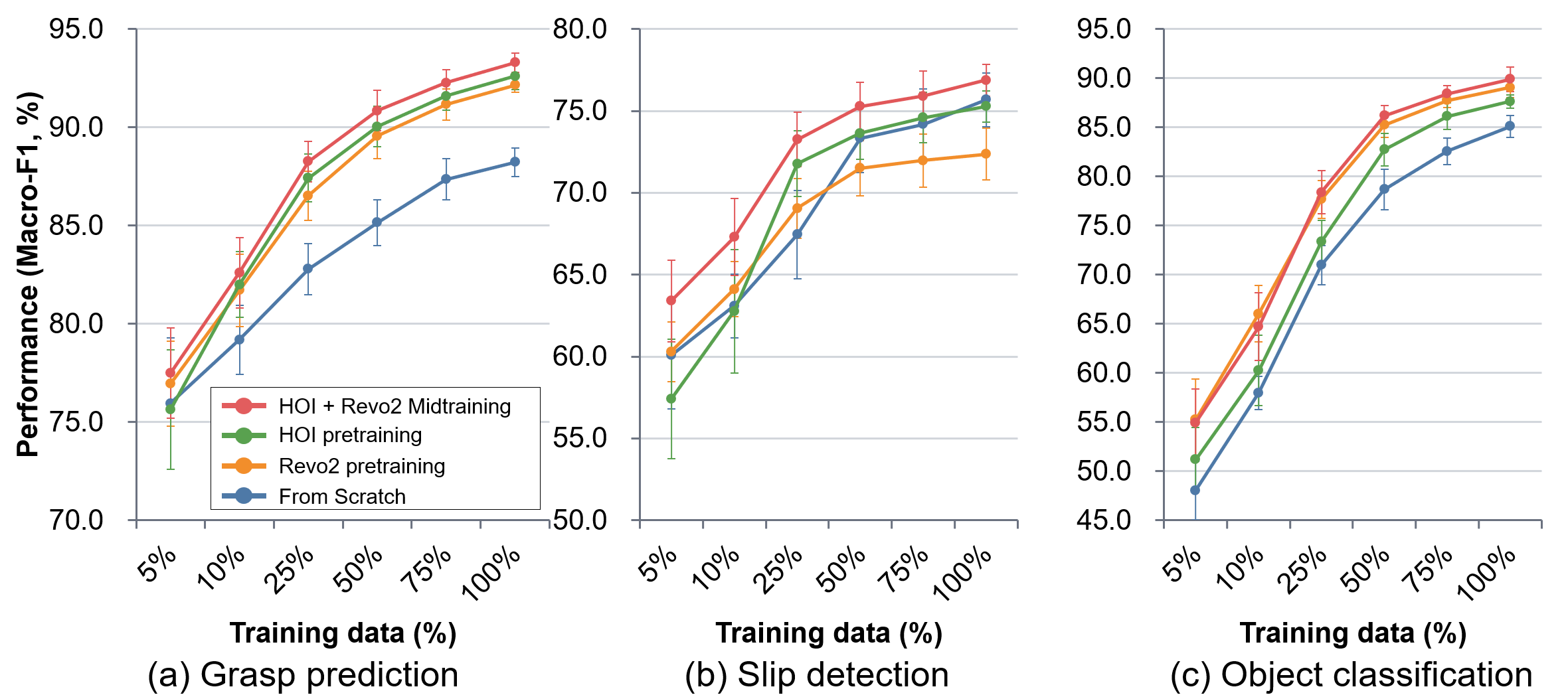}
    \caption{Downstream data efficiency on (a) grasp prediction, (b) slip detection, and (c) object classification. Macro-F1 (\%) is reported as the downstream training data varies from 5\% to 100\%. Error bars denote standard deviation.}
    \label{fig:efficiency}
\end{figure}

We further evaluate the data efficiency of HuTact by varying the amount of downstream training data from 5\% to 100\%. As shown in Fig.~\ref{fig:efficiency}, HOI + Revo2 midtraining consistently reduces the amount of task-specific data required across all three tasks. It matches or exceeds the full-data scratch baseline using only 25\% of the data for grasp prediction (88.3\% vs. 88.2\%), 75\% for slip detection (75.9\% vs. 75.7\%), and 50\% for object classification (86.2\% vs. 85.1\%). These results demonstrate that HuTact substantially improves downstream data efficiency.

\section{Limitations}
Our work has two main limitations. First, geometry-derived pseudo-contact signals do not capture physical contact properties and provide only modest additional gains beyond fingertip coordinates for grasp prediction and object classification. They also require reconstructed hand and object meshes, limiting applicability to HOI datasets with 3D geometry. Second, our evaluation is limited to a Revo2-based bimanual setup, leaving transfer to other robotic hands and tactile sensing modalities unexplored. Future work will investigate richer contact cues from visual and hand-motion observations and extend HuTact to broader robotic embodiments and sensor layouts.
\section{Conclusion}
In this work, we presented HuTact, a framework for pretraining tactile-aware robot hand representations from large-scale human hand-object interaction (HOI) data without requiring real tactile signals during initial pretraining. HuTact leverages fingertip coordinates and pseudo-contact signals, with separate coordinate and tactile tokenization to support transfer to robotic tactile sensing. Across perception and policy-learning tasks, HOI pretraining provides a useful initialization that is further improved by Revo2 midtraining, while also substantially improving downstream data efficiency. These results show that human interaction data can serve as an effective pretraining prior for robotic tactile learning and reduce the need for robot-side tactile data collection.









\bibliographystyle{IEEEtran}
\bibliography{references}

\end{document}